\def\year{2021}\relax
\documentclass[letterpaper]{article}
\usepackage{aaai21}
\usepackage{times}
\usepackage{helvet}
\usepackage{courier}
\usepackage[hyphens]{url}
\usepackage{graphicx}
\usepackage{natbib}
\usepackage{caption}
\usepackage{amsfonts}
\usepackage{amsmath}
\usepackage{amssymb}
\usepackage{cleveref}
\usepackage{nicefrac}
\usepackage{pgfplots}
\usepackage{filecontents}
\usepackage{std_definitions}
\usepackage{multirow}
\usepackage{arydshln}

\title{Have you tried Neural Topic Models? Comparative Analysis of Neural and Non-Neural Topic Models with Application to COVID-19 Twitter Data}
\author {
    Andrew Bennett, \textsuperscript{\rm 1}
    Dipendra Misra, \textsuperscript{\rm 2}
    Nga Than, \textsuperscript{\rm 3} \\
}
 \affiliations {
     \textsuperscript{\rm 1} Department of Computer Science, Cornell University, NY, US. \\
     \textsuperscript{\rm 2} Microsoft Research, New York, US. \\
     \textsuperscript{\rm 3} CUNY Graduate Center, New York, US. \\
     \footnote{Author list is ordered alphabetically. Email addresses of authors are: awb222@cornell.edu, dimisra@microsoft.com, and nthan@gradcenter.cuny.edu}
 }

\newcommand{\ETM}{\mbox{ETM}}
\newcommand{\HDP}{\mbox{HDP}}
\newcommand{\WtoV}{\mbox{W2V}}

\newcommand{\Diversity}{\mbox{TD}}
\newcommand{\softmax}{{\tt softmax}}

\newcommand{\perplexity}{{\tt Perp}}
\newcommand{\coherence}{{\tt Coh}}
\newcommand{\topicgap}{{\tt TG}}
\newcommand{\contrast}{{\tt CS}}

\begin{document}
\maketitle
\begin{abstract}
Topic models are widely used in studying social phenomena. We conduct a comparative study examining state-of-the-art neural versus non-neural topic models, performing a rigorous quantitative and qualitative assessment on a dataset of tweets about the COVID-19 pandemic. Our results show that not only do neural topic models outperform their classical counterparts on standard evaluation metrics, but they also produce more coherent topics, 
which are of great benefit when studying complex social problems. We also propose a novel regularization term for neural topic models, which is designed to address the well-documented problem of mode collapse, and demonstrate its effectiveness.
\end{abstract}

\section{Introduction}

Topic models are routinely used in social sciences to study large unlabeled text corpora~\cite{abebe2019using, roberts2013structural, dimaggio2013exploiting}. However, almost all these approaches use non-neural topic models such as Latent Dirichlet Allocation (LDA), Structural Topic Modeling (STM) or Hierarchical Dirichlet Process (HDP). Recently, neural topic models have been proposed that utilize
word- and topic-embeddings for low-rank modeling of the topic-word probability distributions, which can allow for more flexible control of model complexity as well as the ability to leverage existing methods for computing word embeddings, both of which can lead to improved performance~\cite{miao2017discovering,dieng2020topic}.
This begs the question: how do these approaches perform quantitatively and qualitatively when applied to problems in social sciences? In this paper, we perform a rigorous comparative analysis of a state-of-the-art neural topic model and HDP. 

We study topic modeling on a dataset of tweets about the COVID-19 pandemic. This dataset is large-scale, and is of particular interest from a social science perspective. We first perform a quantitative analysis, where we evaluate various topic models on three automated evaluation metrics, and one human evaluation metric, along the lines of those used in prior topic modelling research. Then, we perform a qualitative analysis, investigating the quality of the learned topics, as well as their usefulness in performing downstream analysis such as measuring the change in topic usage over time. This enables us to study trends in public discourse, as the COVID-19 pandemic spread in early 2020.

Our results show that the base neural topic model suffers from the \emph{mode collapse} phenomenon, which results in repetitive topics. Previous work recommends using word embeddings trained on the dataset, which alleviates this issue to some extent~\cite{dieng2020topic}. We propose a novel differentiable regularization term to further mitigate this phenomenon, and improve topic diversity. We demonstrate that using this regularization term along with word embeddings gives the best topic modeling performance.

We also observe that, while HDP and neural topic models often agree in their topical trends at a high level, the latter tends to produce topics that are more coherent, and provide higher coverage. This can help social scientists extract richer information from their corpora.

\section{Related Work}

There is a rich history of work on topic modelling, dating back to seminal works such as probabilistic Latent Semantic Analysis (pLSA) \citep{hofmann1999probabilistic}, and Latent Dirichlet Allocation (LDA) \citep{blei2003latent}. In particular, LDA has proven to be a popular topic model, since it provides a full generative model that allows the topic distribution of unseen documents to be inferred. Since then, there has a long line of work on constructing different ``LDA-like'' topic models, which modify this standard approach, for example by automatically learning the number of topics to use \citep{teh2006hierarchical}, modelling how topics change over time \citep{blei2006dynamic}, modelling class labels \citep{mcauliffe2007supervised,ramage2011partially}, modelling document metadata or other structural information \citep{mimno2012topic,lee2020incorporating}, modelling structure between topics \citep{griffiths2004hierarchical,titov2008modeling}. As discussed earlier, recently, topic models using neural approaches have also been proposed~\citep{miao2017discovering,dieng2020topic}. In addition, there is a line of work that studies different approaches for fitting such topic models. Originally, \citet{blei2003latent} proposed to fit their model using mean field variational inference. However, since then other approaches have been suggested that have more appealing properties, such as collapsed Gibbs sampling \citep{griffiths2004finding}, which has the advantage of very accurate parameter estimation and inference, or variational inference based on variational autoencoding \citep{kingma2013auto,srivastava2016neural}, which has the advantage of allowing instant inference of unseen documents without further training. 

LDA, and LDA-like methods have been applied widely in textual analysis. However, when the corpora are comprised of a large number of short documents such as social media posts, and tweets, these methods are limited \citep{aldous2019view}. Social scientists, who are trained in discourse analysis, and pay attention to linguistic nuances, have also pointed out that these off-the-shelf methods can only serve as shallow reading of the data, and are not sufficient when the goal is to produce fine-grained categories  \citep{rodriguez2020computational, nelson2018future}. Developing topic models that produce topics with high coherence and coverage is, therefore, an important task.

\section{Topic Modelling Methodology}

In this section, we describe our base topic models models: Hierarchical Dirichlet Process ($\HDP$) and Embedded Topic Model ($\ETM$). Each of these methods takes some vocabulary $\Vcal$ (of size $V$) and a text collection over this vocabulary as input, and returns a topic model which is given by a tuple $(\Tcal,\texttt{infer})$, where $\Tcal = \{T_1, T_2, \ldots, T_K\}$ is a set of $K$ topics, with each topic defined by a distribution over $\Vcal$, and $\texttt{infer}$ is a function that takes a document as input and returns a distribution over $\Tcal$, which define the inferred topic weights for that document. Note that $K$ may be either fixed as a hyperparameter, or learnt by the topic model. \\

\noindent\textbf{HDP.} 
Hierarchical Dirichlet Process \citep{teh2006hierarchical} is a topic model based on a generative model for a text collection, which automatically learns the value of $K$. Formally, the generative model is based on a hierarchy of Dirichlet processes \citep{ferguson1973bayesian}, which is a stochastic process parameterized by a base distribution and concentration parameter, whose sample paths are given by discrete probability distributions over the support of the base distribution. Specifically, it generates the set of topics $\Tcal$ according to a top-level Dirichlet Process whose base distribution is a Dirichlet distribution over the vocabulary, and then independently generates each document according to a Dirichlet Process whose base distribution is given by the mixture of topics generated by the top-level Dirichlet process. This model is fit by collapsed Gibbs sampling \citep{griffiths2004finding}, with updates calculated based on the Chinese restaurant process \citep{aldous1985exchangeability} formalization of the Dirichlet process. Although the top-level Dirichlet process theoretically generates an infinite-set of topics, in practice only a finite number of these are ever used during Gibbs sampling, which is how the model automatically decides the value of $K$ and the set of topics $\Tcal$. Finally, the \texttt{infer} function works by ``folding in'' the document to the model, performing additional iterations of Gibbs sampling on that document only in order to infer its distribution over topics. \\

\noindent\textbf{ETM.} Embedded Topic Model \citep{dieng2020topic} is a topic model based on a low-rank approximation of the Latent Dirichlet Allocation (LDA) topic model, replacing the Dirichlet prior for the document-topic distributions with a Logistic Gaussian \citep{atchison1980logistic} prior in order to facilitate efficient training using variational autoencoders \citep{kingma2013auto}. Specifically, let $\beta \in \mathbb R^{K \times V}$ denote the topic-word probability matrix, whose $(i,j)$'th entry denotes the probability of the $j$'th word in $\Vcal$ in topic $T_i$. The ETM model parameterizes this matrix according to $\beta = \softmax(t v^T)$, where $t \in \mathbb R^{K \times H}$ is a matrix of topic embeddings, $v \in \mathbb R^{V \times H}$ is a matrix of word embeddings, and the hyperparameter $H$ is the embedding dimension. The embeddings matrices $v$ and $t$ are fit using a variational autoencoder algorithm following \citet{srivastava2016neural}, which works by maximizing a lower bound for the log likelihood of the training data known as the Evidence Lower BOund (ELBO). The set of topics $\Tcal$ is then given by these embedding matrices. Note that, unlike HDP, the number of topics $K$ is not automatically inferred, and must be set as a hyperparameter. This variational autoencoder algorithm involves also fitting a neural network $q$ which maps a document to a probability distribution over topics, which at convergence maps a document to its posterior distribution over topics, given the other model parameters $v$ and $t$. Therefore, we can implement the \texttt{infer} function by simply applying the fitted neural network $q$ to the input document. \\

\noindent\textbf{Word2vec Pretraining} \citet{dieng2020topic} also proposed a variation of this model, where the word-embedding matrix was initialized using word2vec embeddings trained on the dataset. In our implementation of this variation, we use the Word2Vec implementation provided by Gensim~\cite{rehurek_lrec}, and continue to fine-tune word embeddings while training the topic model. Note that we do not use any external datasets to train our word embeddings. We call this model $\ETM + \WtoV$. \\

\noindent\textbf{Topic Diversity Regularization.} A known challenge of generative models such as $\ETM$ that are trained via variational autoencoders is \emph{mode collapse}, where the fitted model maps different topics to very similar distributions over words, which occurs due to bad local minima. To avoid this, we propose a diversity regularization term $J(\beta)$, which we define according to\vspace*{-0.1cm}
\begin{equation*}
    J(\beta) = \frac{1}{|\pi|}\sum_{1\le i \le K} \text{TV}(\beta_i, \beta_{\pi(i)}) \,,
\end{equation*}
where $\beta_i$ denotes the $i$'th row of $\beta$, which corresponds to topic $T_i$, $\text{TV}$ denotes total variation distance, $\pi$ is a random permutation of $\{1, 2, \cdots, K\}$, and $|\pi| = \sum_{1\le i\le K} \one\{i \ne \pi(i)\}$. We regularize the ETM model by adding the term $\lambda J(\beta)$ to the ELBO objective to be maximized, where $\lambda \geq 0$ is a hyperparameter controlling the strength of this regularization. We refer to the model that uses this regularizationa term and word2vec as $\ETM + \WtoV + \Diversity$.

\section{COVID-19 Twitter Data}

\begin{table}
\centering
\begin{tabular}{|l|c|}
\hline
\textbf{Dataset Statistics} & \textbf{Values} \\
\hline
\hline
Number of train tweets & 874,975\\
Number of test tweets & 97,285\\
Number of users & 970,816 \\
Vocabulary size & 21,471\\
Number of tokens per tweet & 8.86\\
\hline
\end{tabular}
\caption{Statistcs for the preprocessed Covid-19 dataset containing tweets between Jan 22nd, 2020 and April 30th, 2020.}
\label{tbl:dataset-statistics}
\end{table}

We study topic modeling on a dataset of tweets about the COVID-19 pandemic, provided by~\citet{chen2020tracking}. For computational reasons, we limit the scope of our study to 99 days of data from January 22, 2020 to April 30, 2020.\footnote{With the exception of Feb 23, the day for which data was unavailable in the original corpus.} These 99 days include discussion of the early to mid stages of the COVID-19 pandemic, spanning the time from when COVID-19 was mostly limited to China to when it became a global pandemic. For this purpose, we believe the study of this period will provide crucial insights into public commentary of the pandemic. \\

\noindent\textbf{Preprocessing.} Our corpus consists of 15,156,897 tweets written by 5,049,470 users. For each day in our corpus, we sampled 10,000 tweets without replacement.\footnote{For the first few days in the corpus there were fewer than 10,000 tweets, so for these days we used all available tweets.} We sampled 10\% of tweets without replacement from the dataset to be used as a held-out test split for evaluation, and the remaining tweets (train split) were used for training the model. 

For all tweets in each splits, we performed the following sequence of preprocessing splits: (1) lower case the text; (2) tokenize tweet using the NLTK Twitter Tokenizer; (3) lemmatize each token using NLTK WordNet-based lemmatizer; (4) filter out every token with less than 3 characters; (5) filter out all stop words.\footnote{Based on a custom list we used of 713 stop words.} Finally, we removed any tweets which has no tokens left after these steps. We report aggregate statistics for the processed dataset in Table~\pref{tbl:dataset-statistics}.

\setlength{\tabcolsep}{-1pt}
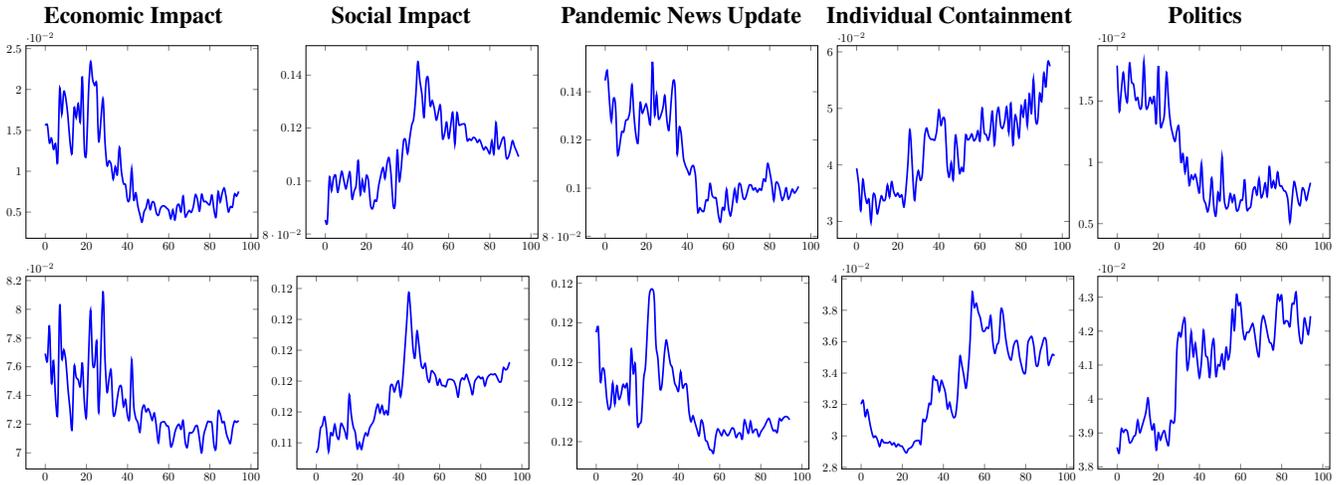
\begin{figure*}
\begin{tabular}{ccccc}
{\footnotesize \textbf{Economic Impact}} & {\footnotesize \textbf{Social Impact}} & {\footnotesize \textbf{Pandemic News Update}} & {\footnotesize \textbf{Individual Containment}} & {\footnotesize \textbf{Politics}}\\
\begin{tikzpicture}[scale=0.45]
\begin{axis}
\addplot[line width=0.5mm,smooth,blue] table [x=x, y=y, col sep=comma] {hdp_economic.csv};
\end{axis}
\end{tikzpicture} &
\begin{tikzpicture}[scale=0.45]
\begin{axis}
\addplot[line width=0.5mm,smooth,blue] table [x=x, y=y, col sep=comma] {hdp_social.csv};
\end{axis}
\end{tikzpicture} &
\begin{tikzpicture}[scale=0.45]
\begin{axis}
\addplot[line width=0.5mm,smooth,blue] table [x=x, y=y, col sep=comma] {hdp_news.csv};
\end{axis}
\end{tikzpicture} &
\begin{tikzpicture}[scale=0.45]
\begin{axis}
\addplot[line width=0.5mm,smooth,blue] table [x=x, y=y, col sep=comma] {hdp_individual_containment.csv};
\end{axis}
\end{tikzpicture} &
\begin{tikzpicture}[scale=0.45]
\begin{axis}
\addplot[line width=0.5mm,smooth,blue] table [x=x, y=y, col sep=comma] {hdp_politics.csv};
\end{axis}
\end{tikzpicture}\\
\begin{tikzpicture}[scale=0.45]
\begin{axis}
\addplot[line width=0.5mm,smooth,blue] table [x=x, y=y, col sep=comma] {etm_economic.csv};
\end{axis}
\end{tikzpicture} &
\begin{tikzpicture}[scale=0.45]
\begin{axis}
\addplot[line width=0.5mm,smooth,blue] table [x=x, y=y, col sep=comma] {etm_social.csv};
\end{axis}
\end{tikzpicture} &
\begin{tikzpicture}[scale=0.45]
\begin{axis}
\addplot[line width=0.5mm,smooth,blue] table [x=x, y=y, col sep=comma] {etm_news.csv};
\end{axis}
\end{tikzpicture} &
\begin{tikzpicture}[scale=0.45]
\begin{axis}
\addplot[line width=0.5mm,smooth,blue] table [x=x, y=y, col sep=comma] {etm_individual_containment.csv};
\end{axis}
\end{tikzpicture} &
\begin{tikzpicture}[scale=0.45]
\begin{axis}
\addplot[line width=0.5mm,smooth,blue] table [x=x, y=y, col sep=comma] {etm_politics.csv};
\end{axis}
\end{tikzpicture}
\end{tabular}
\caption{Topical trend across time for $\HDP$ (top row) and $\ETM+\WtoV+\Diversity$ (bottom row) for five meta-topics. X-axis shows the number of day since Jan 22nd. Y-axis shows the mean probability of meta-topic for documents on a given day.}
\label{fig:topical_analysis}
\end{figure*}

\section{Quantitative Experiments}

First we describe our quantitative experiments on the COVID-19 Twitter Dataset. For these experiments, we compare the previously described topic models using several automated or human evaluation metrics, described below.

\subsection*{Evaluation Metrics} 

\noindent\textbf{Perplexity ($\perplexity$):} this metric is computed according to $\exp(-\ln(L) / n_{\text{tok}})$, where $\ln(L)$ is the estimated log likelihood of the held out test data according to the given topic model, and $n_{\text{tok}}$ is the total number of tokens in the test data. This is based on the assumption that a good topic model will predict a high likelihood for held-out documents, and therefore will have low perplexity. \\

\noindent\textbf{Coherence ($\coherence$):} For any given words $w$ and $w'$, let $P_{\text{test}}(w)$ and $P_{\text{test}}(w,w')$ respectively denote the probability of $w$ appearing in a randomly sampled test document, and the probability of $w$ and $w'$ co-occurring in a randomly sampled test document, and let $\text{npmi}(w,w')$ denote the normalized pointwise mutual information between $w$ and $w'$, defined according to
\begin{equation*}
    \text{npmi}(w,w') = \frac{\log \frac{P_{\text{test}}(w,w')}{P_{\text{test}}(w) P_{\text{test}}(w')}}{- \log P_{\text{test}}(w,w')}\,.
\end{equation*}
In addition, let $w_i^{(k)}$ be the $i$'th highest-probability word in the $k$'th topic. Then, the coherence metric is calculated according to $\frac{1}{K} \sum_{k=1}^K \frac{1}{45} \sum_{i=1}^{10} \sum_{j=i+1}^{10} \text{npmi}(w_i^{(k)}, w_j^{(k)})$, where $K$ is the number of topics. This is based on the assumption that the top words of a topic should co-occur often, and therefore a good topic model will have high coherence. \\

\noindent\textbf{Topic Gap ($\topicgap$):} This metric measures \emph{diversity} between different topics. To compute this, we take the union of top 10 highest-probability words from each of the $K$ topics, and compute the metric according to $n_{\text{unique}} / 10K$, where $n_{\text{unique}}$ is the number of unique words in this union. If each topic generates a disjoint set of top 10 words, then the topic gap will have a maximum value of 1. A higher value, denotes less repetitive topics, which is a desirable property. \\

\noindent\textbf{Human Evaluation ($\contrast$):} This metric is motivated by past work on topic and word intrusion tests~\citep{chang2009reading,schnabel2015evaluation}. We used the following setup for our human study: annotators were provided with pairs of word lists, each containing five words sampled from the same topic. To generate these lists, we first sample a topic $k$ from its prior probability. Then, we uniformly sample 5 words from the set of top 10 highest probability words for this topic, and create our first list $\Ucal_1$ using them. With 50\% probability, we use the remaining five words as the second list $\Ucal_2$. Otherwise, we randomly sample a different topic $l$, and we create the list $\Ucal_2$ by uniformly sampling 5 words from $l$'s top 10 highest probability words (after removing words in $\Ucal_1$). Annotators \footnote{All authors of this paper equally annotated the samples.}  were then asked to predict whether $\Ucal_1$ and $\Ucal_2$ were sampled from the same or different topics. For each method, we annotated 100 such pairs in total.\footnote{Under the constraint that we only select a given topic $k$ once.} For all methods, we then calculated a \emph{contrast score}, which we define as the fraction of correct annotations for that method. Ideally, this metric should capture similar information to the automated coherence and topic gap metrics, since in order for correct annotation to be possible the top words in each topic must be at least somewhat cohesive and distinct from the top words in other topics. However, unlike these automated metrics it has the advantage that it can leverage linguistic intuitions. \\

\setlength{\tabcolsep}{4pt}
\begin{table}[t]
    \centering

    \begin{tabular}{lcccc}
        \hline
        \textbf{Topic Model} & \textbf{$\perplexity$} & \textbf{$\coherence$} & \textbf{$\topicgap$} & \textbf{$\contrast$} \\
        \hline
        $\HDP$ &  7875.4 & 0.03 & 0.39 & 0.81 \\
        $\ETM$ &  2831.1 & 0.05 & 0.12 & 0.73\\
        $\ETM + \WtoV$ &  2754.9 & \textbf{0.11} & 0.21 & 0.86\\
        $\ETM + \WtoV + \Diversity$ & \textbf{2312.5} & 0.09 & \textbf{0.55} & \textbf{0.88}\\
        \hline
    \end{tabular}
    \caption{Performance of topic models on different metrics.}
    
    \label{tab:metric-results}
\end{table}

\noindent\textbf{Implementation Details.} In the case of $\HDP$, we used the tomotopy implementation. \footnote{https://bab2min.github.io/tomotopy/} We trained the model for 4,000 iterations, and evaluated it every 100 iterations. For $\ETM$, we used the code provided by the authors.\footnote{https://github.com/adjidieng/ETM} We trained the model for 100 epochs using batch sizes of 1000, and evaluated every 2,500 iterations. For every experiment, we use perplexity on the test set to select model.

For all methods, we performed grid search over hyperparameters. For $\HDP$, this included the three hyperparameters ($\alpha$, $\beta$, $\eta$) that control the Dirichlet processes, and for $\ETM$, it included number of topics, learning rate, hidden dimension, all word2vec hyperparameters, and the weight of topic diversity regularizer. We selected the best hyperparameters using the three automated metrics.\footnote{This was done by considering all metrics, with hyperparameter configuration $A$ prefered over $B$ if $A$ outperformed $B$ on more metrics than vice versa.} \\

\noindent\textbf{Results.} We report performance of topic models in Table~\ref{tab:metric-results}. We observe that $\HDP$ generates reasonably diverse topics, but these topics lack coherence and have high perplexity on held-out test data. In comparison, all $\ETM$ models achieve significantly lower perplexity and higher coherence. We speculate that this may be due to 
the low-rank factorization of the topic-word distribution employed by neural topic models, which reduces the effective number of parameters and may help prevent over-fitting.
The base model $\ETM$, however, suffers from \emph{mode collapse} and does not generate a diverse set of topics. It also receives the lowest contrast score, indicating that the annotators were unable to differentiate topics. Using word2vec initialization ($\ETM+\WtoV$) significantly improves all four metrics. However, the topic gap remains below $\HDP$. Lastly, our proposed variant $\ETM + \WtoV + \Diversity$ outperforms other models on most metrics, and in particular, achieves the highest human evaluation score.

\begin{table*}
{\fontsize{6.5}{10}\selectfont
    \centering
    \begin{tabular}{p{0.49\textwidth}p{0.49\textwidth}}
  \hline
 \textbf{Politics} & \textbf{Social Impact} \\
  \hline \hline
  \textbf{$\HDP$} & \\
  \hline
  china, coronavirus, \#coronavirus, chinese, outbreak, minister, covid, president, health, fight & coronavirus, covid, church, lockdown, \#coronavirus, service, pastor, china, pandemic, \#covid19 \\
   coronavirus, vote, election, trump, pandemic, biden, covid, bernie, voting, voter & corona, coronavirus, virus, china, covid, cancelled, due, outbreak, hope, week \\
   coronavirus, boris, lockdown, johnson, covid, government, \#coronavirus, minister, news, \#covid19 & coronavirus, league, game, china, covid, player, season, corona, football, team \\
   china, trump, money, deal, trade, biden, american, coronavirus, billion, ukraine & covid, coronavirus, \#covid19, service, home, lockdown, \#coronavirus, pandemic, stay, due \\
   china, communist, country, russia, trump, america, coronavirus, party, bernie, korea & school, coronavirus, student, covid, class, online, university, home, \#coronavirus, china \\
   \hline
   \textbf{$\ETM + \WtoV$} & \\
   \hline
  trump, president, democrat, vote, election, republican, penny, hoax, donald, biden & event, cancelled, cancel, due, canceled, sport, 2020, postponed, player, league \\
   trump, american, medium, fact, lie, truth, president, racist, cdc, blame & today, video, watch, show, free, online, join, love, game, friend \\
   2020, march, april, feb, february, jan, january, refund, ticket, due & school, close, area, order, open, student, closed, city, border, shut \\
   china, country, chinese, america, war, usa, deal, trade, power, citizen & \#stayhome, \#stayathome, \#lockdown, \#staysafe, \#socialdistancing, \#quarantine, \#covid19, \#quarantinelife, \#stayhomesavelives, \#corona \\
   china, chinese, country, communist, usa, america, war, party, russia, ccp & covid, lockdown, week, home, family, due, place, month, call, today \\
   \hline
   \textbf{$\ETM + \WtoV + \Diversity$} & \\
   \hline
  house, county, white, governor, gov, boris, california, florida, york, johnson & due, event, game, cancelled, cancel, concern, trip, postponed, ticket, fan \\
   government, law, act, action, policy, federal, court, failed, legal, nigerian & pandemic, crisis, plan, working, fight, part, community, response, team, hard \\
   trump, president, american, america, democrat, lie, administration, vote, blame, obama & family, friend, love, hope, guy, feel, happy, message, kind, hey \\
   health, public, official, emergency, national, minister, organization, general, authority, security & week, today, school, due, order, lockdown, hour, class, move, return \\
   party, war, communist, power, ccp, police, political, mass, china's, protest & city, quarantine, place, close, open, shut, area, border, closed, local \\
   \hline
    \end{tabular}}
    \caption{We present the top 10 words for 5 randomly sampled topics falling under two different meta-topics, for our different topic-modelling methods.}
    \label{tab:top-words}
\end{table*}

\section{Topic Analysis of COVID-19 Twitter Data}

We finally present our qualitative analysis, using the topics produced by our $\HDP$ and $\ETM$ topic models on the COVID-19 Twitter dataset. The goals of this analysis are twofold: we wish to understand the topical trends in this data, as well as compare the extent to which we can perform such analysis with our different topic models. At a high level, our analysis proceeds as follows: (1) we run our best-performing $\HDP$ and $\ETM$ topic models from the previous experiments on the  dataset; (2) we extract the topics from these, and cluster them into 11 meta-topics, each with multiple sub-topics;\footnote{This was done based on the topics' top words, and by ``deep reading" 50 exemplar tweets per topic \citep{nelson2020computational}.} and (3) we calculate the prevalence of these meta-topics over time according to our two topic models.

Based on our analysis of the topics produced by these models, we decided on the following meta-topics: (1) \emph{China}; (2) \emph{Economic Impact}; (3) \emph{Social Impact}; (4) \emph{Politics}; (5) \emph{Individual Containment Measures}; (6) \emph{Administrative Response}; (7) \emph{Frustration and Anger}; (8) \emph{Hospitals and Healthcare}; (9); \emph{Pandemic News Updates}; (10) \emph{Information about the Virus}; and (11) \emph{Misinformation}. In addition, we used an additional \emph{Miscellaneous} meta-topic for any tweet that did not fit into our core meta-topics. Sub-topics contain finer information. For example, \emph{fear}, \emph{prayer}, and \emph{frustration with administration} are sub-topics of the meta-topic \emph{frustration}.

First, we plot the prevalence over time of five of these meta-topics in Figure~\ref{fig:topical_analysis}, for both our HDP and the best ETM model. We can see here that in most cases, the general trend of HDP and ETM agree with each other. For example, for the \emph{Social Impact} and \emph{Pandemic News Update} plots the trends predicted by the two models are almost identical. Furthermore, for the \emph{Economic Impact} and \emph{Individual Containment} topics, although the trends aren't quite as identical, they are broadly very similar. Conversely, in the case of \emph{Politics} the trends are very different.

We believe that one major factor explaining the above difference is that the topics produced by HDP, as summarized by top words, tend to be less cohesive and more difficult to interpret compared to those from ETM. In practice, this may lead to more noisy meta-topic labelling for HDP, and more topics labelled as miscellaneous, especially in more challenging edge cases. Furthermore, this intuition is backed up by the fact that of the five meta-topics in Figure~\ref{fig:topical_analysis}, the \emph{Politics} meta-topic where they disagree is a very broad topic with many edge-cases related to other meta-topics (such as \emph{Administrative Response} and \emph{China}), and therefore is more prone to this issue. Note that this finding is consistent with past work (\emph{e.g.} \citet{dieng2020topic}), which find that low-rank embedding-based models such as ETM tend to produce higher quality topics compared to more classical models such as HDP. 

Next, we directly examine the quality of the topics produced by these different topic models on this dataset. In Table~\ref{tab:top-words} we present the top 10  words for 5 randomly sampled topics from the \emph{Politics} and \emph{Social Impact} meta-topics. We can make a few immediate observations from these topics. First, we note that the $\HDP$ topics are particularly noisy; its topic word lists are littered with words that are common to the entire dataset but not to any more specific topic (such as ``coronavirus'' or ``\#covid19''). Second, the best performing ETM model ($\ETM + \WtoV + \Diversity$) has many topics that are extremely clean and specific; for example, within the \emph{Social Impact} meta-topic it has a topic very specifically about quarantine and lockdowns, a topic very specifically about cancelled events, a topic very specifically about the community resopnse, \emph{etc}. In comparison, the HDP topics seem to be generally much more vague; for example again within the \emph{Social Impact} meta-topic many of the topics are hard to pin down very specifically, beyond being about the social impact of COVID-19. Third, although it seems very clear that ETM is tending to produce topics that are cleaner and more coherent, it is difficult to compare the coverage of the topics given that the HDP topics are typically much more noisy, so it is difficult to judge how many aspects of each meta-topic are covered by HDP. Nonetheless, we argue qualitatively that ETM appears to be achieving good coverage, with topics that cover many different specific aspects of each meta-topic, which is consistent with its high topic gap score presented previously. Finally, we also included results for the second best performing ETM model(($\ETM + \WtoV$) in Table~\ref{tab:top-words} for comparison. We note that these topics seem to be relatively clean and high quality in comparison with those from HDP (reflecting high coherence), but do not seem to cover as many specific aspects of each meta-topic (reflecting relatively low topic gap). In particular, this observation about lower coverage again reinforces the mode-collapse challenges of ETM, and the importance of our topic diversity regularization.

\section*{Conclusion} Our study suggests that neural topic modeling is beneficial for studying complex social issues. While some argue that topic modeling in social science can only serve as the first level of shallow human coding, our research shows that with the new development of neural topic modeling, analysts can extract more interpretable and richer categories from their corpora.

\paragraph{Acknowledgement.} Our study was reviewed and approved by the Microsoft Research Institutional Review Board (IRB). We thank the IRB reviewers for their help and feedback. We would also like to thank Alexandra Olteanu, Maria Antoniak, and Alexandra Schofield for valuable discussions. Finally, our study would not have been possible without computational resources and support provided by the Microsoft GCR team.

\paragraph{Reproducibility.} Code for reproducing our results can be found at \url{https://github.com/ngathan/etm-covid-analysis}. Please see~\citet{chen2020tracking} for information about the dataset.

\bibliographystyle{aaai21}
\bibliography{references}
\end{document}